\documentclass[letterpaper, 10 pt, conference]{ieeeconf}  % Comment this line out if you need a4 paper

\IEEEoverridecommandlockouts                              % This command is only needed if
\usepackage{mleftright}
\usepackage{algorithm}
\usepackage{pbox}
\usepackage{algpseudocode}
\usepackage{setspace,epsfig,psfig,amssymb,amsfonts,arydshln,graphicx,subeqnarray,amsmath,epstopdf, color, comment, multirow, bm}
\usepackage{psfrag}

\usepackage{amsthm}

\title{\LARGE \bf Fast Online Planning for Bipedal Locomotion via Centroidal Model Predictive Gait Synthesis }

\author{Yijie Guo$^1$, Mingwei Zhang$^1$, Hao Dong$^1$, Mingguo Zhao$^2$% <-this % stops a space
\thanks{$^{1}$Yijie Guo,  Mingwei Zhang and Hao Dong are with Beijing Research Institute of UBTECH Robotics, Beijing, China.
{\tt\small 	$\{$fugo.guo, mingwei.zhang, howard.dong$\}$@ubtrobot.com}}
\thanks{$^{2}$Mingguo Zhao is with the Department of Automation, Tsinghua University, Beijing, China.
{\tt\small mgzhao@mail.tsinghua.edu.cn}}
}

\begin{document}
\maketitle
\thispagestyle{empty}
\pagestyle{empty}

%%%%%%%%%%%%%%%%%%%%%%%%%%%%%%%%%%%%%%%%%%%%%%%%%%%%%%%%%%%%%%%%%%%%%%%%%%%%%%%%
\begin{abstract}
The planning of whole-body motion and step time for bipedal locomotion is constructed as a model predictive control (MPC) problem, in which a sequence of optimization problems needs to be solved online. While directly solving these problems is extremely time-consuming, we propose a predictive gait synthesizer to offer immediate solutions. Based on the full-dimensional model, a library of gaits with different speeds and periods is first constructed offline. Then the proposed gait synthesizer generates real-time gaits at 1kHz by synthesizing the gait library based on the online prediction of centroidal dynamics. We prove that the constructed MPC problem can ensure the uniform ultimate boundedness (UUB) of the CoM states and show that our proposed gait synthesizer can provide feasible solutions to the MPC optimization problems. Simulation and experimental results on a bipedal robot with 8 degrees of freedom (DoF) are provided to show the performance and robustness of this approach.
\end{abstract}

%%%%%%%%%%%%%%%%%%%%%%%%%%%%%%%%%%%%%%%%%%%%%%%%%%%%%%%%%%%%%%%%%%%%%%%%%%%%%%%%
\section{INTRODUCTION}
Bipedal robots are complex dynamic systems with high degrees of freedom. Different approaches have been investigated for real-time motion planning for bipedal robots. Classical methods based on reduced-order models have been well developed\cite{raibert1986legged, kajita2014introduction, englsberger2015three}, while the workspace of robots may be limited and some physical constraints (actuator bounds, friction cone, etc.) are not directly considered. Thus, a lot of recent work  \cite{dai2014whole, posa2016optimization,dafarra2020whole} focused on trajectory optimization based on full-dimensional models. However, due to the complexity and non-convexity of the formulated nonlinear optimization problems, these methods are not yet ready for online implementation. 

In order to consider full-dimensional dynamics/constraints and avoid online trajectory optimization, gait library based methods have been proposed. Through offline trajectory optimization based on the full-dimensional model, these methods first construct a library of periodic or aperiodic gait trajectories, then choose the appropriate trajectory online. One of the earliest work based on this idea is \cite{westervelt2003switching}, where gaits with fixed speeds are designed for a planar underactuated biped. This idea was later extended to fully-actuated bipedal robots in \cite{powell2013speed, murali2019optimal}. In these studies, the gait trajectory is chosen according to the speed command. The stability of the robot heavily relies on the controller, since the motion planning does not change according to robot states. Under large disturbances, these methods may fail when the controller can not track the planned motion due to the physical constraints of actuators or ground reaction forces. To address this issue, a gait updating method is proposed in \cite{da20162d}, the gait trajectory is updated according to the mid-step average speed, thus the robots can handle larger speed perturbations. However, as the gait is updated only once at the middle of each step, it can not react in time for disturbances near the beginning or end of each step.  

In the meantime, current gait library methods generally keep a constant step time, while adjusting both step location and step time greatly enlarges the margin of stability \cite{feng2016online}. It is shown in \cite{koolen2012capturability} that a shorter step time results in a larger capturability region. However walking consistently with a very short step time is unnatural and power-consuming, and users may require a specific step time in some cases. This brings the need for online step time adjustment to meet the user command under normal circumstances and ensure stability under disturbances.

In this paper, we propose a gait synthesis approach from an MPC point of view. The proposed gait synthesizer generates real-time gaits by synthesizing a multi-period gait library based on the online prediction of centroidal dynamics. This enables fast reactive gait updating at 1kHz, and the step time is adjusted online by synthesizing gaits with different periods. There is also related work in \cite{wang2017robust,khadiv2020walking,scianca2020mpc,smaldone2021feasibility,daneshmand2021variable} on planning step timing and location using MPC techniques. They generally use the LIP model to construct an online solvable MPC problem, subject to the stability/viability condition proposed also based on the LIP model. While our approach constructs a whole-body MPC problem that considers the whole-body constraints. Instead of directly solving this MPC problem as people usually do, we find immediate solutions by gait library synthesizing, which avoids the extremely time-consuming online solving process. More importantly, we prove that the constructed MPC problem can ensure the UUB stability of the CoM states and show that our proposed gait synthesizer can provide feasible solutions to the MPC optimization problems. At last, simulation and experimental results show that robots can achieve versatile and robust locomotion with this proposed approach.

The paper is organized as follows. The MPC problem for locomotion planning is described in Sec. II. Then the multi-period gait library is constructed in Sec. III. In Sec. IV, the predictive gait synthesizer is proposed and shown to offer feasible solutions to the MPC optimization problems, the UUB stability of the post-impact CoM states is also proved in this section. Simulation and experimental results on an 8-DoF bipedal robot are presented in Sec. V. Finally, conclusions and future directions are given in Sec. VI. 

\section{Problem Description}
\begin{figure}[htb]
   \begin{center}
    \includegraphics[width=0.485\textwidth]{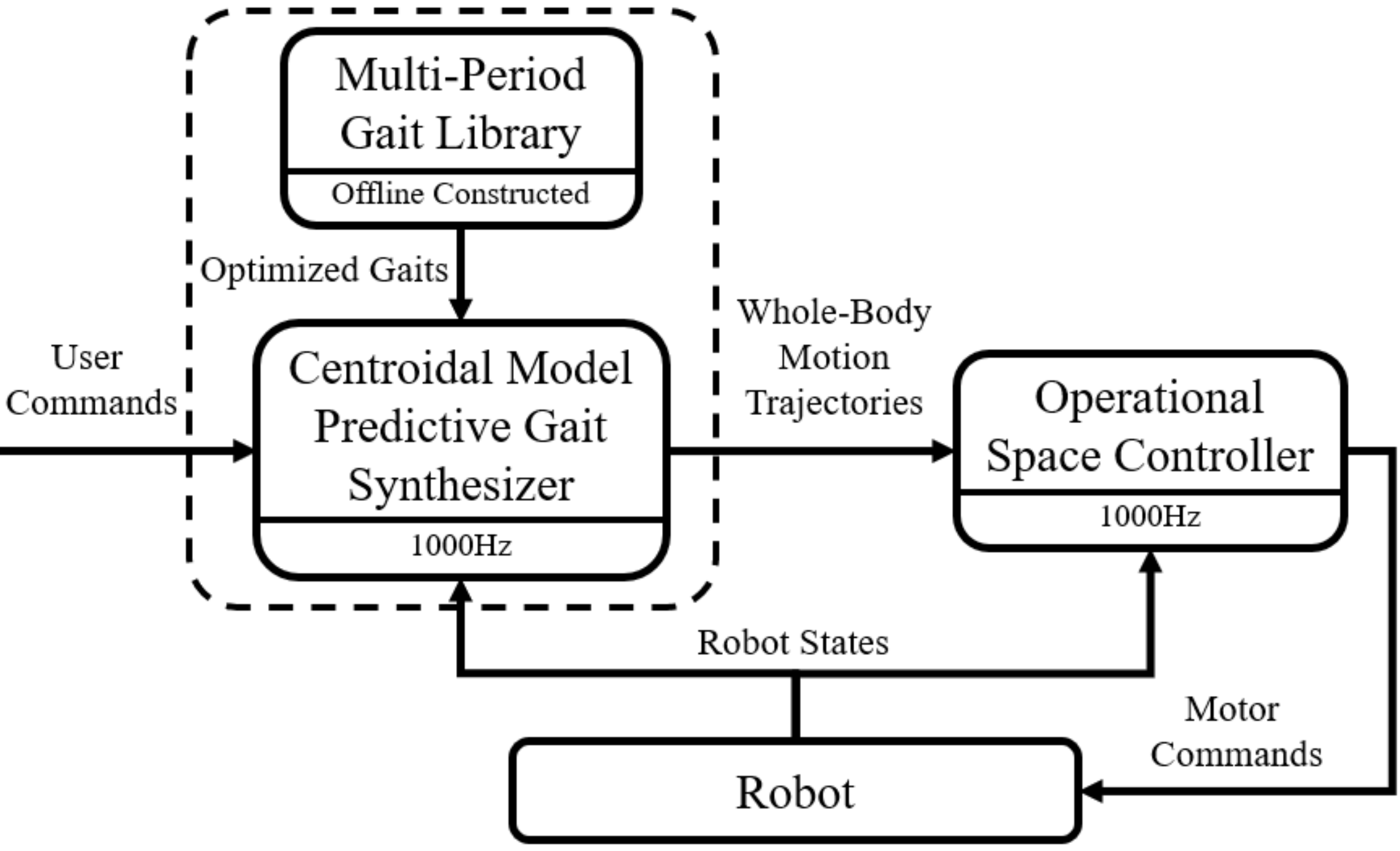}
   \caption{The overall motion planning and control architecture. The proposed centroidal model predictive gait synthesizer is shown in the dashed box.}
      \label{pic:system_overview}
   \end{center}
\end{figure}

The overall motion planning and control architecture is shown in Fig. \ref{pic:system_overview}. The planner generates whole-body motion trajectories according to the user command and current robot states, then an operational space controller (OSC) generates appropriate motor commands to follow these trajectories. We focus on the motion planning part in this paper.

Our proposed gait synthesizer is an MPC style planner, the basic idea is to solve the following optimization problem for future 2 steps at $t_0$, the time from the last foot impact (assume the right leg is the stance leg at the $i^{\text{th}}$ step):
\begin{equation}\label{eq:mpc}
\begin{aligned}
& \underset{\varphi(t), T_t^i,T_t^{i+1}}{\text{min}}
& & \int_{t_0}^{T_t^i+T_t^{i+1}} ||\tau_{\varphi(t)} \cdot \omega_{\varphi(t)}||^2 dt\\
\end{aligned}
\end{equation}
\begin{equation*}
\begin{aligned}
& \text{s.t.}
& &||v^+_x[i+2]-v^+_{x,d}|| \leq k_1||v^+_x[i+1]-v^+_{x,d}|| + \epsilon_x\\
& & & ||v^+_y[i+2]-{}_Rv^+_{y,d}|| \leq k_2||v^+_y[i+1]-{}_Lv^+_{y,d}|| + \epsilon_y\\
& & & ||p^+_x[i+2]-p^+_{x,d}|| \leq k_3\Big|\Big|
\begin{bmatrix}
    v^+_x[i+2]\\
    v^+_y[i+2]\\
\end{bmatrix}-
\begin{bmatrix}
    v^+_{x,d}\\
    {}_Rv^+_{y,d}\\
\end{bmatrix}
\Big|\Big|\\
& & & ||p^+_y[i+2]-{}_Rp^+_{y,d}|| \leq k_4\Big|\Big|
\begin{bmatrix}
    v^+_x[i+2]\\
    v^+_y[i+2]\\
\end{bmatrix}-
\begin{bmatrix}
    v^+_{x,d}\\
    {}_Rv^+_{y,d}\\
\end{bmatrix}
\Big|\Big|\\
& & & \text{Joint position, velocity and torque constraints satisfied.}\\
& & & \text{Ground reaction force (GRF) constraints satisfied.}
\end{aligned}
\end{equation*}
where $\varphi(t)$ represents whole-body motion trajectories, $T_t^i$ is the step duration, $\tau_{\varphi(t)}$ and $\omega_{\varphi(t)}$ are the joint torques and velocities to achieve $\varphi(t)$, $p^+[i]=[{p}^{+}_x[i],{p}^{+}_y[i]]^T$ and $v^+[i]=[v^{+}_x[i],v^{+}_y[i]]^T$ are the post-impact horizontal CoM position and velocity relative to the stance foot, $p^+_d=[{p}^+_{x,d},{}_{R/L}{p}^+_{y,d}]^T$ and $v^+_d=[v^+_{x,d},{}_{R/L}{v}^+_{y,d}]^T$ are the desired values, subscripts $x,y$ indicate the x/y direction, subscripts $R,L$ indicate the right/left leg is the stance leg. The first four constraints are stability constraints, the constants $k_1,k_2 \in(0,1)$ and $k_3,k_4 \in(0,\infty)$ ensures the uniform ultimate boundedness (UUB) of $v^+$ and $p^+$ around $v^+_d$ and $p^+_d$, as shown in Sec. IV.C. The bounded set is determined by the small non-negative constants $\epsilon_x$ and $\epsilon_y$. With $\epsilon_x$ and $\epsilon_y$ being 0, $v^+$ and $p^+$ will be exponentially converged to $v^+_d$ and $p^+_d$.

However, directly solving (\ref{eq:mpc}) using whole-body trajectory optimization is extremely time-consuming. In the following sections, we show that our proposed gait synthesizer can provide a feasible solution by combining an offline constructed multi-period gait library and online gait synthesis based on centroidal dynamics. As shown in Fig. \ref{pic:system_overview}, this gait synthesizer can run at 1kHz, the same as the OSC. This fast re-planning greatly increases the robot robustness to disturbances and environmental uncertainties. 

\section{Multi-period Gait Library} \label{sec:PMGT}

In this section, the multi-period gait library is first constructed through trajectory optimization based on the full-dimensional model.

\subsection{Hybrid Model of Walking}
The walking process is modeled as a hybrid system, including a single support phase and an instantaneous double support phase. Assuming the right leg is the stance leg, the overall hybrid model of walking can be written as:
\begin{equation}\label{eq:hybird_model}
\left\{\begin{array}{cc}
    \begin{array}{c}D(q)\ddot{q} + H(q,\dot{q}) = B\tau + J_R(q)^Tf_R\\
    J_R(q)\ddot{q} + \dot{J}_R(q,\dot{q})\dot{q} = 0\vspace{1.5ex}\end{array}
    &(q,\dot{q})\notin S\\
    \dot{q}^+ = \Delta(q) \dot{q}^-&(q,\dot{q})\in S
\end{array}\right.
\end{equation}
where $S=\{(q,\dot{q})|\,p^z_L(q)=0,\, \dot{p}^z_L(q,\dot{q})\leq0\}$.

The first two equations in (\ref{eq:hybird_model}) describe the single support phase dynamics. The first equation describes the floating base dynamics, where $q$ is the vector of generalized coordinates including both floating states and joint states, $D(q)$ is the mass-inertia matrix, $H(q,\dot{q})$ contains the gravity force and coriolis force, $\tau$ is the vector of motor torques, $f_R$ is the contact wrench. $B$ is the motor torque distribution matrix, $J_R(q)$ is the Jacobian matrix of the contact point. The second equation describes the contact constraint, i.e., the acceleration of the contact point is zero. 

The third equation in (\ref{eq:hybird_model}) describes the instantaneous double support phase. When the vertical position of the swing foot $p^z_L(q)$ decreases to 0, i.e. $(q,\dot{q})\in S$, the robot enters double support phase. Following the rigid impact development process in \cite{westervelt2018feedback}, the double support phase can be modeled as a discrete map $\Delta(q)$ between $\dot{q}^-$ and $\dot{q}^+$ (the velocity of the system just before and after impact).

\subsection{CoM Related Outputs}
Each gait in the gait library contains time trajectories of the selected quantities to be controlled, these quantities are termed as 'outputs'. In existing work, joint angles are usually selected as outputs \cite{westervelt2003switching, da20162d} for direct use of joint-level control. Recently, workspace quantities are also used for their intuitive physical meanings \cite{gong2019feedback}, which can then be combined with an OSC. In this paper, CoM related quantities are selected as outputs to support the CoM based gait synthesis in Sec. IV.

Specifically, 10 quantities are selected as outputs, they are listed in Tab. \ref{tab:output}. The CoM x and y positions are not selected as outputs as the stance foot roll and pitch are in fact weakly actuated for limited sole area, these two actuations are used in the controller to regulate the pre-impact CoM velocity for robots with active feet. For underactuated bipedal robots, certain quantities are removed from the outputs. For example, for a bipedal robot with fully-passive feet, $\phi_{foot}$, $\theta_{foot}$ should be removed. 

\begin{table}[htb]
    \caption{Selected Outputs}\label{tab:output}
	\centering
	\resizebox{0.4\textwidth}{!}{$\begin{tabular}{l|l}
        \hline
		Torso roll, pitch and yaw & $\phi_{torso}$, $\theta_{torso}$, $\psi_{torso}$ \\      
        \hline
        Vertical position of the CoM & $z_{COM}$ \\
        \hline
        Swing foot x and y positions & $x_{foot}$ \\
        relative to the CoM & $y_{foot}$ \\
        \hline
        Swing foot vertical position & $z_{foot}$ \\
        \hline
        Swing foot roll, pitch and yaw & $\phi_{foot}$, $\theta_{foot}$, $\psi_{foot}$ \\
        \hline
    \end{tabular}$}
\end{table}

\subsection{Periodic Gaits Optimization}
For a periodic gait, the sagittal pre-impact CoM velocity ($v^-_x$) repeats every step, while the lateral pre-impact CoM velocity (${}_Rv^-_y$ and ${}_Lv^-_y$) repeats every two steps. Therefore, in order to cover different periodic sagittal and lateral movements, we need a gait library with at least four dimensions $[T, v^-_x, {}_Rv^-_y, {}_Lv^-_y]$. As the development of trajectory optimization tools such as FROST\cite{hereid2017frost} and C-FROST \cite{hereid2019rapid}, this gait library can be constructed conveniently.  

Periodic walking gaits with different sagittal, lateral average velocities and different lateral velocity differences are optimized for multiple periods. The difference of lateral velocities $\delta v_{y}^- = {}_Rv_{y}^- - {}_Lv_{y}^-$ is also constrained as different combinations of ${}_Rv_{y}^-$ and ${}_Lv_{y}^-$ can realize the same average lateral velocity. Each optimization problem is performed over two steps with the right and left leg being the stance leg successively. The cost function used in the optimization is the sum square of power:
\begin{equation}\label{eq:COST}
\text{Cost} = \int_0^{2T} ||\tau \cdot \omega||^2 dt,\\
\end{equation}
where $\tau$ and $\omega$ are the actuated joint torques and velocities.
Constraints enforced in the optimization are listed in Tab. \ref{tab:gait constr}. The swing foot impact velocity is constrained to 0 so that the foot impact will not change the CoM velocity, i.e.
\begin{equation}\label{eq:v_impact_const}
v^+_x = v^-_x, v^+_y = v^-_y.
\end{equation}

\begin{table}[htb]
    \caption{Constraints used in gait optimization}\label{tab:gait constr}
	\centering
	\resizebox{0.324\textwidth}{!}{$\begin{tabular}{l|l}
        \hline
		Average Sagittal Velcoity & $\bar{v}_{x,i}$ \\      
        \hline
        Average Lateral Velcoity & $\bar{v}_{y,i}$ \\
        \hline
        Difference of Lateral Velocity  & $\delta v_{y,i}^-$ \\
        \hline
        Period & $T_i$ \\
        \hline
        Friction Cone & $\mu = 0.6$ \\
        \hline
        Mid-step Swing Foot Height & $0.07m$ \\
        \hline
        Swing Foot Impact Velocity & $(0, 0, 0)m/s$\\
        \hline
        Joint Position, Velocity and & Determined \\
        Torque Limits & by hardware \\
        \hline
    \end{tabular}$}
\end{table}

After these optimizations, we acquire output trajectories for different gaits. All these trajectories are parameterized with B\'ezier polynomials in terms of the normalized time $s=t/T\in[0,1]$. The $i^{\text{th}}$ output $h_d^i$ can be represented as 
\begin{equation}\label{eq:bezier}
\begin{array}{l}
    h_d^i(s) = \sum_{j=1}^{M+1}{\alpha}(i,j)\frac{M!}{(j-1)!(M-j+1)!}s^{j-1}(1-s)^{M-j+1},\\
\end{array}
\end{equation}
where $M$ is the order of the B\'ezier polynomial, $N$ is the number of outputs, $\alpha \in \mathbb{R}^{N \times (M+1)}$ is the parameter matrix. Thus, each gait can be represented with a parameter matrix labeled with its unique $[T, v^-_x, {}_Rv^-_y, {}_Lv^-_y]$, i.e., $\alpha^T_{ v^-_x, {}_Rv^-_y, {}_Lv^-_y}$.

\textit{Remark 1:} A simplified gait library can be constructed with 2 dimensions $[T, v^-_x]$, the nominal lateral footstrike location can be calculated online using the LIP model \cite{xiong2019orbit}:
\begin{equation}\label{eq:y_foot}
y_{foot}(T) = (v_{y}^--d)/{\sigma},\\
\end{equation}
where $\sigma = \lambda tanh(\frac{T}{2}\lambda)$, $\lambda = \sqrt{\frac{g}{\bar{z}_{COM}}}$,
$d=\frac{\lambda^2 sech(\frac{T}{2}\lambda)T}{2\sigma}\bar{v}_{y}$, $T$ is the current period, $\bar{z}_{COM}$ is the average COM height of current gait, $\bar{v}_y$ is the desired average lateral velocity .

  \textit{Remark 2:} Standing motion can be considered as a periodic trajectory with an $\infty$ period. The posture at the beginning of a gait with zero average sagittal and lateral velocity can be directly used for standing.

\section{Centroidal Model Predictive Gait Synthesizer}

\begin{comment}
\begin{figure}[htb]
   \begin{center}
    \includegraphics[width=0.34\textwidth]{mpgs.pdf}
   \caption{Our proposed gait synthesizer first predicts $\dot{p}^-[i]$, $p^-[i]$, then synthesizes a gait that drives $\dot{p}^+[i+2]$, $p^+[i+2]$ towards $\dot{p}^{+*},p^{+*}$.}
      \label{pic:mpgs}
   \end{center}
\end{figure}
\end{comment}
In this section, we propose the MPC style gait synthesizer. It provides a feasible solution of (\ref{eq:mpc}) by first predicting the pre-impact CoM states $v^-[i]$, $p^-[i]$ according to current $v, p$ and then synthesizing gaits accordingly.

\subsection{Pre-Impact CoM States Prediction}
The pre-impact CoM states $v^-$, $p^-$ are first predicted according to current $v, p$. The centroidal dynamics of a robot can be described as:
\begin{equation}\label{eq:IP_model}
\left\{
\begin{array}{ll}
    \ddot{z} = \dfrac{1}{m}f_z - g\vspace{1ex}\\
    \ddot{p} = \dfrac{1}{m}f_p,
\end{array}
\right.
\end{equation}
where $z$ is the vertical position of the CoM, $p$ represents the horizontal position of the CoM, which can be either x or y axis position, $f_p$ and $f_z$ are ground reaction forces, $m$ is the total mass, $g$ is the acceleration of gravity. It is assumed that there is enough rotational friction between the foot and the ground to balance the z axis angular momentum. The x/y axis angular momentum change rate is also assumed to be 0, as the swing and stance legs are generally symmetric and the torso is kept upright, i.e.
\begin{equation}\label{eq:ang_moment}
f_pz-f_zp=0\\
\end{equation}
Combining (\ref{eq:IP_model}) and (\ref{eq:ang_moment}), we can have
\begin{equation}\label{eq:predict_model}
\begin{bmatrix}
    \ddot{z}\\
    \ddot{p}\\
\end{bmatrix}
=
\begin{bmatrix}
    \dfrac{1}{m}\vspace{1ex} & -1\\
    \dfrac{p}{mz} & 0\\
\end{bmatrix}
\begin{bmatrix}
    f_z\\
    g\\
\end{bmatrix}, 
\end{equation}
where $f_z$ is determined by the PD controller in the OSC,
\begin{equation}\label{eq:fz}
    f_z = (\ddot{z}^* + k_p(z^* - z) + k_d(\dot{z}^*-\dot{z}) + g)m,
\end{equation}
where $z^*, \dot{z}^*, \ddot{z}^*$ are current CoM vertical reference trajectory and its derivatives, $k_p$ and $k_d$ are controller parameters (As the robot used in this paper is a joint-torque-controlled robot and the OSC controller explicitly considers the full-dimensional dynamics, the OSC can accurately regulate the vertical acceleration and hence $f_z$ during stance). Rearranging (\ref{eq:predict_model}) and (\ref{eq:fz}), we can have a nonlinear state space model
\begin{equation}
    \dot{X} = f_{predict}(X,z^*, \dot{z}^*, \ddot{z}^*,t),
\end{equation}
where $X=[z,p,\dot{z},v]^T$. Given current state $X(t_0)$, the pre-impact states can be predicted by numerical integration,
\begin{equation}\label{eq:predict_model2}
    X^- = X(T_t) = \int_{t_0}^{T_t} f_{predict}(X,z^*, \dot{z}^*, \ddot{z}^*,t)dt.
\end{equation}

For standing gait, the CoM height is kept constant, the Capture Point (CP) is calculated instead of predicting the CoM states at $T_t=\infty$. As the CP is the nearest zero momentum point (ZMP) that can reduce the horizontal CoM velocity to 0, the support region should at least contain the CP for the robot to keep standing. Thus the CP is used as the criterion to predict if the robot can keep standing. The CP is calculated by
\begin{equation}\label{eq:capture_point}
        CP = p + v/\sqrt{g/z}.\\
\end{equation}

Note that the CoM height is assumed constant for standing to simplify the control architecture, other more advanced standing controllers \cite{koolen2016balance} that can vary the CoM height can also be used in this gait synthesis framework.

\subsection{Gait Synthesis Algorithm}
\begin{algorithm}[htb]\label{al:1}
\caption{Gait Synthesis Algorithm}\label{euclid}
\begin{algorithmic}[1]
\State \textbf{Input:} $p$, $v$, $t_0$, $s_0$, $j_d$, $v^+_d$, $i$
\State \textbf{Output:} $\alpha$
\State \textbf{Initialize:} $Flag=0$
\For{$j = j_d,j_d+1,\cdots,k$}
\State $T=S_T\{j\}$, $T_t = t_0 + (1-s_0)T$
\State \textbf{Predict} $p^-[i], v^-[i]$ using $p$, $v$, $t_0$, $T_t$ with (\ref{eq:predict_model2})
\If{$(p^-[i], v^-[i])\in S_{feasible}$}
\State \textbf{Gait Interpolation:}
\begin{equation}\label{eq:gait_intp}
\begin{array}{ll}
     & v^-_{y,s}[i] = {}_Rv^+_{y,d} + {}_Lv^+_{y,d} - v^-_y[i], \\
     & \alpha = G_{intp}(T,v^-_x[i],v^-_y[i],v^-_{y,s}[i]).
\end{array}
\end{equation}
\State \textbf{Gait Modification:}
\begin{equation}\label{eq:gait_modi}
\begin{array}{ll}
     & \alpha(5,M+1) = \alpha(5,M+1) + k_x(v^-_x[i] - v^+_{x,d}),  \\
     & \alpha(6,M+1)= \alpha(6,M+1) - k_y(v^-_{y}[i] - {}_Lv^+_{y,d}).\\
     & \text{Then check kinematic constraints.}
\end{array}
\end{equation}
\State $Flag = 1$, \textbf{Break;}
\EndIf
\EndFor
\If{$Flag ==0$}
\State Prepare for falling
\EndIf
\end{algorithmic}
\end{algorithm}

Assume the gait library has $k$ periods, $l$ sagittal pre-impact velocities and $n$ pair of lateral pre-impact velocities. The set of periods is $S_T = \{T_1, T_2, \cdots, T_k\}$ in descending order. The sets of pre-impact velocities are $S_{v^-_{x}} = \{v^-_{x,1}, v^-_{x,2}, \cdots, v^-_{x,l}\}$, $S_{{}_Rv^-_{y}} = \{{}_Rv^-_{y,1}, {}_Rv^-_{y,2}, \cdots, {}_Rv^-_{y,n}\}$ and $S_{{}_Lv^-_{y}} = \{{}_Lv^-_{y,1}, {}_Lv^-_{y,2}, \cdots, {}_Lv^-_{y,n}\}$ in which velocities are in ascending order. From now on, the right leg is assumed to be the stance leg, subscripts $R$ and $L$ can be swapped for the left leg case. The gait synthesis algorithm is summarized by the pseudo code in Algorithm 1.

In lines 1-3 the input, output and initialization of the algorithm are defined. The input includes current robot states $p, v$, the time from the last foot impact $t_0$, the normalized step time $s_0$, the index $j_d$ of the desired period $T_d$($T_d = S_T\{j_d\}$) and the desired post-impact velocities $v^+_d$, $i$ is the current step number. The output is the synthesized gait parameter $\alpha$. The variable $Flag$ is the flag for finding a feasible $\alpha$.

In lines 4-7 the period set $S_T$ is traversed from the $j_d$-th element, until a feasible $T$ is found. And the total step duration $T_t$ is modified accordingly. If the period is changed to $T$ at $t_0$ (normalized step time $s_0$), then $s$ is updated with the new period $T$ by $s = s + \frac{t_s}{T}$ to keep $s$ continuous, until $s=1$, where $t_s$ is the sample time. Thus the total step duration of this step is $T_t = t_0 + (1-s_0)T$. A $T$ is feasible if the predicted $(p^-[i], v^-[i])$ with this $T$ is in $S_{feasible}$. The feasible set $S_{feasible}$ is defined as
\begin{equation}
\{(p^-, v^-)|p^-\in S_{CoM}, g_{foot}(T,v^-_x,v^-_y,v^-_{y,s}) \in S_{foot}\},
\end{equation}
where $g_{foot}(T,v^-_x,v^-_y,v^-_{y,s})$ (part of $G_{intp}$ in (\ref{eq:intp_func})) calculates the pre-impact swing foot position of the interpolated gait, $S_{CoM}$ and $S_{foot}$ are feasible regions of the CoM and swing foot positions, they are designed to satisfy the kinematic limits and avoid foot collision. If the predicted $(p^-[i], v^-[i])$ with current $T$ is not feasible, a smaller $T$ is tested as decreasing step time enlarges the feasible region.
  
In line 8 the gait parameter $\alpha$ is synthesized if a feasible period $T$ is found. First, the gait library is interpolated by tri-linear interpolation: 
\begin{equation}\label{eq:intp_func}
\begin{array}{l}
G_{intp}(T,v^-_x[i],v^-_y[i],v^-_{y,s}[i])=\\
+\alpha^T_{v^-_{x,u},{}_Rv^-_{y,v},{}_Lv^-_{y,w}}(1-\xi_1)(1-\xi_2)(1-\xi_3)\\
+\alpha^T_{v^-_{x,u+1},{}_Rv^-_{y,v},{}_Lv^-_{y,w}}\xi_1(1-\xi_2)(1-\xi_3)\\
+\alpha^T_{v^-_{x,u},{}_Rv^-_{y,v+1},{}_Lv^-_{y,w}}(1-\xi_1)\xi_2(1-\xi_3)\\
+\alpha^T_{v^-_{x,u},{}_Rv^-_{y,v},{}_Lv^-_{y,w+1}}(1-\xi_1)(1-\xi_2)\xi_3\\
+\alpha^T_{v^-_{x,u+1},{}_Rv^-_{y,v},{}_Lv^-_{y,w+1}}\xi_1(1-\xi_2)\xi_3\\
+\alpha^T_{v^-_{x,u+1},{}_Rv^-_{y,v+1},{}_Lv^-_{y,w}}\xi_1\xi_2(1-\xi_3)\\
+\alpha^T_{v^-_{x,u},{}_Rv^-_{y,v+1},{}_Lv^-_{y,w+1}}(1-\xi_1)\xi_2\xi_3\\
%\end{array}
%\end{equation}
%\begin{equation*}
%\begin{array}{l}
+\alpha^T_{v^-_{x,u+1},{}_Rv^-_{y,v+1},{}_Lv^-_{y,w+1}}\xi_1\xi_2\xi_3,
\end{array}
\end{equation}
if $v^-_{x,u} \leq v^-_x[i] \leq v^-_{x,u+1}$, ${}_Rv^-_{y,v} \leq v^-_y[i] \leq {}_Rv^-_{y,v+1}$,${}_Lv^-_{y,w} \leq v^-_{y,s}[i] \leq {}_Lv^-_{y,w+1}$, where $\xi_1 = \frac{v^-_x[i] - v^-_{x,u}}{v^-_{x,u+1}- v^-_{x,u}}$, $\xi_2 = \frac{v^-_y[i] - {}_Rv^-_{y,v}}{{}_Rv^-_{y,v+1}- {}_Rv^-_{y,v}}$, $\xi_3 = \frac{v^-_{y,s}[i] - {}_Lv^-_{y,w}}{{}_Lv^-_{y,w+1}-{}_Lv^-_{y,w}}$, $0\leq u<l$, $0\leq v<n$, $0\leq w<n$. This interpolated gait is approximately a periodic gait, which can drive $v^-_x[i+1], v^-_y[i+1]$ to a very small neighbor of $v^-_x[i], v^-_{y,s}[i]$ as explained in (\ref{eq:assump1}). 

In line 9 $\alpha$ is partially modified to regulate $v^-[i+1]$ towards $v^+_d$ with a discrete P-type controller, $\alpha(5,M+1)$ and $\alpha(6,M+1)$ represent the pre-impact swing foot positions $x_{foot}$ and $y_{foot}$ respectively\footnote{$x_{foot}$ and $y_{foot}$ are the 5$^\text{th}$ and 6$^\text{th}$ outputs in Tab. \ref{tab:output}, they are parameterized by the 5$^\text{th}$ and 6$^\text{th}$ rows of $\alpha$, and for a B\'ezier polynomial, the last coefficient equals to the end value, i.e., the pre-impact value.\label{note1}}. This type of controller has been successfully implemented in \cite{raibert1986legged,da20162d, xiong2019orbit,rezazadeh2015spring}. After the gait modification, the kinematic feasibility of the swing foot position is checked again, if violated, the gait modification will be truncated.

In line 10, the traversing is break when a new feasible gait is synthesized. In lines 13-15 the robot prepares for falling if no feasible gait is found.

During standing, if the CP is outside the support region, the robot will use Algorithm 1 to adapt to a finite period. When the CP is within the support region at a double support instant, the robot can switch to standing.

\subsection{Stability Analysis}
In this section, we present the stability analysis of $v^+$ and $p^+$ in terms of UUB stability. First the solution of (\ref{eq:mpc}) is shown to ensure the UUB stability of $v^+$ and $p^+$.
\newtheorem*{theorem1}{Theorem 1.1}
\begin{theorem1}
The solution of (\ref{eq:mpc}) ensures the uniform ultimate boundedness of $v^+$ around $v^+_d$, i.e., there $\exists b,c>0$, for every $0<a<c$, there exists $N = N(a,b) \in \mathbb{N}_+$ such that
\begin{equation}
    ||v^+[i_0]-v^+_d||\leq a \Rightarrow  ||v^+[i]-v^+_d|| \leq b, \forall i\geq i_0+N
\end{equation}
\end{theorem1}
\emph{Proof:} We prove for $v^+_x$ here, the proof for $v^+_y$ is similar.
Let $c = min(||v^-_{x,1}-v^+_{x,d}||, ||v^-_{x,n}-v^+_{x,d}||)$, i.e., the minimal distance from $v^+_{x,d}$ to the velocity boundary of the gait library. Let $b>\frac{\epsilon_x}{1-k_1}$, and $N \geq log_{k_1}\frac{b-\frac{\epsilon_x}{1-k_1}}{a}$.
Then according to the first constraint in (\ref{eq:mpc}), and $k_1 \in (0,1)$, we can have
\begin{equation}
\begin{array}{ll}
    ||v^+[i]-v^+_d|| \leq {k_1}^{i-i_0}||v^+[i_0]-v^+_d|| + \sum_{j=0}^{i-i_0-1} k_1^j\epsilon_x\\
    \leq {k_1}^{i-i_0}a + \frac{1-k_1^{i-i_0}}{1-k_1}\epsilon_x \leq {k_1}^{i-i_0}a + \frac{\epsilon_x}{1-k_1}.
\end{array}
\end{equation}
Since $i-i_0\geq N \geq log_{k_1}\frac{b-\frac{\epsilon_x}{1-k_1}}{a}$ and $k_1 \in (0,1)$ 
\begin{equation}
\begin{array}{ll}
    ||v^+[i]-v^+_d||\leq {k_1}^Na + \frac{\epsilon_x}{1-k_1} \leq \frac{b-\frac{\epsilon_x}{1-k_1}}{a}a + \frac{\epsilon_x}{1-k_1} = b,
\end{array}
\end{equation}
which completes the proof.\hfill $\blacksquare$

\newtheorem*{theorem2}{Theorem 1.2}
\begin{theorem2}
The solution of (\ref{eq:mpc}) also ensures the uniform ultimate boundedness of $p^+$ around $p^+_d$.
\end{theorem2}
\emph{Proof:} Since $v^+$ is uniformly ultimately bounded, according to the 3rd and 4th constraints in (\ref{eq:mpc}), $p^+_x$ and $p^+_y$ are also uniformly ultimately bounded, with the bounded set being $k_3(k_4)$ times the bounded set of $v^+$.\hfill $\blacksquare$

\textit{Remark 3:}
If $\epsilon_x = 0$ and $\epsilon_y = 0$, it can be easily shown that the solution of (\ref{eq:mpc}) ensures the exponential stability of $v^+$ and $p^+$ towards $v^+_d$ and $p^+_d$.

Next we show that the generated gaits of our gait synthesizer satisfy the stability constraints of (\ref{eq:mpc}).  The velocity range of the gait library is defined as
\begin{equation}\label{eq:S_v}
\begin{array}{ll}
    S_v = \{(v^-_x, {}_Rv^-_y, {}_Lv^-_y)|v^-_{x,1}\leq v^-_x \leq v^-_{x,l}, \\
    {}_Rv^-_{y,1} \leq {}_Rv^-_y \leq {}_Rv^-_{y,n},{}_Lv^-_{y,1} \leq {}_Lv^-_y \leq {}_Lv^-_{y,n}\}.
    \end{array}
\end{equation}
The poincar\'e map of the pre-impcat CoM velocities between steps are defined as $P_{map}$, i.e.,
\begin{equation}
        [v_{x}^-[i+1],v_{y}^-[i+1]] = P_{map}(\alpha_i, v_{x}^-[i],v_{y}^-[i]),
\end{equation}
where $\alpha_i$ is the gait parameter implemented for the $i^{\text{th}}$ step.

Then an assumption is given about the gait interpolation.
\newtheorem*{assumption1}{Assumption 1} 
\begin{assumption1}
There exist $\epsilon_x, \epsilon_y \geq 0$ such that, for any $[v^-_x, {}_Rv^-_y, {}_Lv^-_y]\in S_v$ and $\alpha_{intp} = G_{intp}(T,v^-_x, {}_Rv^-_y, {}_Lv^-_y)$, $[P_{map}^x, P_{map}^y] = P_{map}(\alpha_{intp}, v^-_x, {}_Rv^-_y)$ satisfy:
\begin{equation}\label{eq:assump1}
    ||P_{map}^x - v^-_x|| \leq \epsilon_x, \,||P_{map}^y - {}_Lv^-_y|| \leq \epsilon_y.
\end{equation}
\end{assumption1}
\newtheorem*{assumption2}{Assumption 2.2} 

This assumption is based on the fact that each gait $\alpha^T_{v^-_x, {}_Rv^-_y, {}_Lv^-_y}$ in the periodic gait library satisfies
\begin{equation}\label{eq:periodic}
    P_{map}(\alpha^T_{v^-_x, {}_Rv^-_y, {}_Lv^-_y}, v^-_x, {}_Rv^-_y) = [v^-_x, {}_Lv^-_y],
\end{equation}
thus the interpolated gait should approximately satisfy (\ref{eq:periodic}) at the in-between velocities with very small errors $\epsilon_x$ and $\epsilon_y$. They are bounded by the grid size of the gait library, as the grid size decreases, $\epsilon_x$ and $\epsilon_y$ converge to 0.
\newtheorem*{lemma1}{Lemma 2.1} 
\begin{lemma1}
There exist $\delta_x, \delta_y < 0$ such that, the ratios of the pre-impact CoM velocity change $\Delta P_{map}$ to the footstrike location change $\Delta \alpha(5,M+1)$ and $\Delta \alpha(6,M+1)$ satisfy:
\begin{equation}\label{eq:assump2}
\begin{array}{cc}
     \delta_x\leq\frac{\Delta P_{map}^x}{\Delta \alpha(5,M+1)} < 0, \delta_y\leq\frac{\Delta P_{map}^y}{\Delta \alpha(6,M+1)}<0.
\end{array}
\end{equation}
\end{lemma1}

According to (\ref{eq:predict_model}), $\ddot{p} = \frac{f_z}{mz}p$. Then we can discretize it to
\begin{equation}\label{eq:predict_model_discrete}
\begin{bmatrix}
    p(k+1)\\
    v(k+1)\\
\end{bmatrix}
=
\begin{bmatrix}
    1 & \Delta t\\
    \frac{f_z(k)}{mz(k)}\Delta t & 1\\
\end{bmatrix}
\begin{bmatrix}
    p(k)\\
    v(k)\\
\end{bmatrix}.
\end{equation}
For a given nominal gait, $z(k)$ is a fixed trajectory, so is $f_z(k)$. Thus $\frac{f_z(k)}{mz(k)}$ is a fixed trajectory, besides $\frac{f_z(k)}{mz(k)}$ is positive and bounded. Then we can multiply (\ref{eq:predict_model_discrete}) iteratively for a step period $T$ to get the pre-impact CoM velocity, given initial CoM states $p_0$ and $v_0$, and show that the change ratio is negative and bounded. Note that the initial CoM position equals to the negative value of the footstrike location.\hfill$\blacksquare$

\newtheorem*{lemma2}{Lemma 2.2} 
\begin{lemma2}
Each element $g_{intp}(T,v^-_x, {}_Rv^-_y, {}_Lv^-_y)$ of $G_{intp}(T,v^-_x, {}_Rv^-_y, {}_Lv^-_y)$ is Lipschitz continuous in $S_v$, i.e., there $\exists K>0$ such that, for all $V_a = [v^-_{x,a}, {}_Rv^-_{y,a}, {}_Lv^-_{y,a}]$ and $V_b = [v^-_{x,b}, {}_Rv^-_{y,b}, {}_Lv^-_{y,b}]$ in $S_v$,
\begin{equation}\label{eq:Lipschitz}
    ||g_{intp}(T,V_a) - g_{intp}(T,V_b)|| \leq K||V_a - V_b||.
\end{equation}
\end{lemma2}
According to (\ref{eq:intp_func}), every $g_{intp}$ is a cubic function in each subset of $S_v$ and can be shown to be Lipschitz continuous in each subset. Then it can be further shown to be Lipschitz continuous in $S_v$, since $S_v$ is convex. The detailed proof is omitted due to space limit.\hfill $\blacksquare$

\newtheorem*{theorem3}{Theorem 2} 
\begin{theorem3}
The generated gait of our gait synthesizer satisfies the stability (the first four) constraints of (\ref{eq:mpc}), if
\begin{equation}\label{eq:kxky}
\begin{array}{c}
    0 < k_x < -\frac{2}{\delta_x}, 0 < k_y < -\frac{2}{\delta_y}.
\end{array}
\end{equation}
\end{theorem3}
\emph{Proof:} (Assume the right leg is the stance leg for step $i$) 

We here prove for the sagittal direction. First, we show
\begin{equation}\label{eq:condition1}
    ||v^+_x[i+2]-v^+_{x,d}|| \leq k_1||v^+_x[i+1]-v^+_{x,d}|| + \epsilon_x
\end{equation}
According to (\ref{eq:gait_intp}),(\ref{eq:gait_modi}):
\begin{equation}
    \begin{array}{l}
    v^-_x[i+1] = P_{map}^x + \frac{\Delta P_{map}^x}{\Delta \alpha(5,M+1)}k_x(v^-_{x}[i]-v^{+}_{x,d}).
    \end{array}
\end{equation}
Thus 
\begin{equation}
\begin{array}{ll}
    ||v^-_x[i+1]-v^+_{x,d}||\vspace{1ex}\\
     =||P_{map}^x - v^-_{x}[i] + (1+ \frac{\Delta P_{map}^x}{\Delta \alpha(5,M+1)}k_x)(v^-_{x}[i]-v^{+}_{x,d})||\vspace{1ex}\\
     \leq||1+ \frac{\Delta P_{map}^x}{\Delta \alpha(5,M+1)}k_x||||v^-_{x}[i]-v^{+}_{x,d}|| + ||P_{map}^x - v^-_{x}[i]||.
\end{array}
\end{equation}

According to (\ref{eq:assump1}),(\ref{eq:assump2}) and (\ref{eq:kxky}), $||P_{map}^x - v^-_{x}[i]||\leq \epsilon_x$, $||1+ \Delta P_{map}^x / \Delta \alpha(5,M+1) k_x|| \leq ||1+\delta_xk_x||<1$, thus 
\begin{equation}
    ||v^-_x[i+1]-v^+_{x,d}||\leq ||1+\delta_xk_x||||v^-_{x}[i]-v^{+}_{x,d}|| + \epsilon_x.
\end{equation}

Furthermore, (\ref{eq:v_impact_const}) indicates that $v^+[i+1]=v^-[i]$, thus (\ref{eq:condition1}) is proved with $k_1 = ||1+\delta_xk_x||$.
Next we show that 
\begin{equation}\label{eq:condition3}
||p^+_x[i+2]-p^+_{x,d}|| \leq k_3\Big|\Big|
\begin{bmatrix}
    v^+_x[i+2]\\
    v^+_y[i+2]\\
\end{bmatrix}-
\begin{bmatrix}
    v^+_{x,d}\\
    {}_Rv^+_{y,d}\\
\end{bmatrix}
\Big|\Big|\\
\end{equation}

The pre-impact swing foot position is part of the output of $G_{intp}$ (as explained in Sec. IV-B\textsuperscript{\ref{note1}}), noted as $g_{foot}$. The post-impact CoM position equals to the negative value of the pre-imapct swing foot position, i.e., $p^+[i+1]$ = $-p_{foot}^{-}[i]$. Thus for the desired periodic gait, $p^+_d$ and $v^{+}_d$ should satisfy 
\begin{equation}
    p^+_{x,d} = -g^x_{foot}(T,v^+_{x,d}, {_R}v^+_{y,d}, {_L}v^+_{y,d}).
\end{equation}
Consider the $i+1^{\text{th}}$ step, according to (\ref{eq:gait_intp}),(\ref{eq:gait_modi}), 
\begin{equation}
\begin{array}{ll}
     & p^+_x[i+2] = -g_{foot}^x(T, v^-_x[i+1], v^-_{y}[i+1], v^-_{y,s}[i+1])\vspace{1ex}\\
     & - k_x(v^-_x[i+1]-v^+_{x,d})\\
\end{array}
\end{equation}
Thus,
\begin{equation}
    \begin{array}{ll}
    ||p^+_x[i+2]-p^+_{x,d}||\vspace{1ex} = ||g_{foot}^x(T,v^-_{aug}[i+1])\\ +k_x(v^-_x[i+1]-v^+_{x,d}) - g_{foot}^x(T,v^+_{x,d}, {_R}v^+_{y,d}, {_L}v^+_{y,d})||\vspace{1ex}\\
    \leq||g_{foot}^x(T,v_{aug}^-[i+1]) - g_{foot}^x(T,v^+_{x,d}, {_R}v^+_{y,d}, {_L}v^+_{y,d})|| +\vspace{1ex}\\ k_x||v^-_x[i+1]-v^+_{x,d}||
    \end{array}
\end{equation}
where $v^-_{aug}[i+1] = [v^-_x[i+1],v^-_y[i+1], v^-_{y,s}[i+1]]$.

Then, according to (\ref{eq:v_impact_const}) and (\ref{eq:Lipschitz}) 
\begin{equation}
    \begin{array}{ll}
         ||p^+_x[i+2]-p^+_{x,d}||\\
         \leq K\Bigg|\Bigg|
         \begin{bmatrix}
             v^+_x[i+2] - v^+_{x,d}\\
             v^+_y[i+2] - {_R}v^+_{y,d}\\
             v^+_y[i+2] - {_R}v^+_{y,d}\\
         \end{bmatrix}\Bigg|\Bigg| + k_x||v^+_x[i+2]-v^+_{x,d}||\\    
    %\end{array}
%\end{equation}
%\begin{equation*}
    %\begin{array}{l}
         \leq (\sqrt{2}K + k_x)\Big|\Big|
        \begin{bmatrix}
            v^+_x[i+2]\\
            v^+_y[i+2]\\
        \end{bmatrix}-
        \begin{bmatrix}
            v^+_{x,d}\\
            {}_Rv^+_{y,d}\\
        \end{bmatrix}
        \Big|\Big|
    \end{array}
\end{equation}
Thus (\ref{eq:condition3}) is proved with $k_3 = \sqrt{2}K + k_x$. 
The proof for the lateral direction is similar, it is omitted here.\hfill $\blacksquare$
\subsection{Kinematic and Dynamic Feasibility}
The kinematic limits including the joint position and velocity limits are explicitly considered in the offline gait optimization. To further ensure the kinematic feasibility of the interpolated and modified gaits, the feasibility of the CoM and the swing foot position is checked in line 7 in Algorithm 1 during gait interpolation, if violated, a smaller $T$ will be used. Then in line 9, the feasibility of the swing foot position is checked again after gait modification.

The generated gaits are dynamic feasible if the joint torques and ground reaction forces to realize the output trajectories are within feasible limits. These limits are also explicitly considered in the offline gait optimization. Despite the loss of theoretical soundness, this provides a good foundation for the interpolated and slightly modified gaits to also satisfy these feasibility constraints. In practical, we can set more conservative constraints in the offline optimizations and constrain the modified term in (\ref{eq:gait_modi}) to ensure the dynamic feasibility of the generated gaits.

In our simulations and experiments, the constraints in the offline optimizations are not tightened and the modified value in (\ref{eq:gait_modi}) is less than 5\% of the leg length, and the generated gaits are always kinematically and dynamically feasible. 

Overall, the generated gaits of our gait synthesizer satisfy the stability constraints of (\ref{eq:mpc}), and we can use the gait library and the pre-mentioned actions to ensure their kinematic and dynamic feasibility practically. Thus our proposed gait synthesizer can provide feasible solutions to (\ref{eq:mpc}).

\section{Implementation Results}\label{sec:results}

\begin{figure}[htb]
   \begin{center}
    \includegraphics[width=0.36\textwidth]{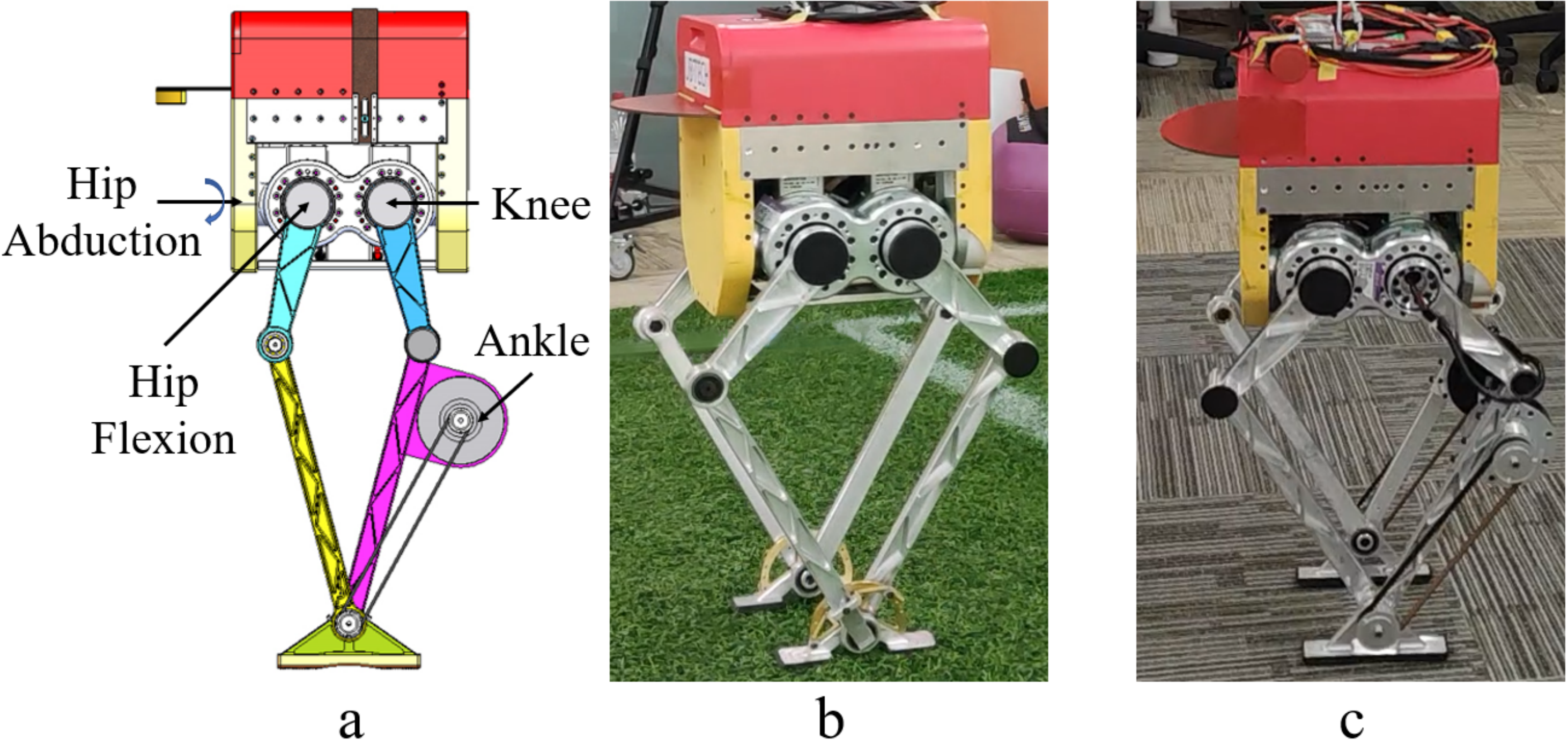}
   \caption{Bipedal robots for simulations and experiments. (a) Simulation model. (b) passive-foot robot. (c) active-foot robot.}
      \label{pic:Mario}
   \end{center}
\end{figure}

This section presents simulation and experimental results of the proposed gait synthesizer. As shown in Fig. \ref{pic:Mario}a, an 8-DoF bipedal robot is used in the simulation, it has four actuators in each leg, they are for hip abduction, hip flexion, knee and ankle respectively. A physical robot is built according to this model first with passive feet (Fig. \ref{pic:Mario}b) and then with active feet (Fig. \ref{pic:Mario}c). The active-foot robot weighs 17 kg and the hip height is 0.41m in standing pose. 

\subsection{Simulation Results}
We first show simulation results on the 8-DoF robot model. The gait library is constructed with 3 periods $\{\infty,\,0.35,\,0.2\}(s)$ and 8 sagittal average velocities $\{-0.5,\,-0.3,\,-0.15,\,0,\,0.15,\,0.3,\,0.5,\,0.7\}(m/s)$, the periodic gait for the lateral direction is calculated using (\ref{eq:y_foot}). The controller used in the simulation is a QP-based operational space controller, similar to the one used in \cite{apgar2018fast}. The gait modification parameters are $k_x=0.08, k_y=0.095$, same for all simulations.

\begin{figure}[htb]
   \begin{center}
    \includegraphics[width=0.48\textwidth]{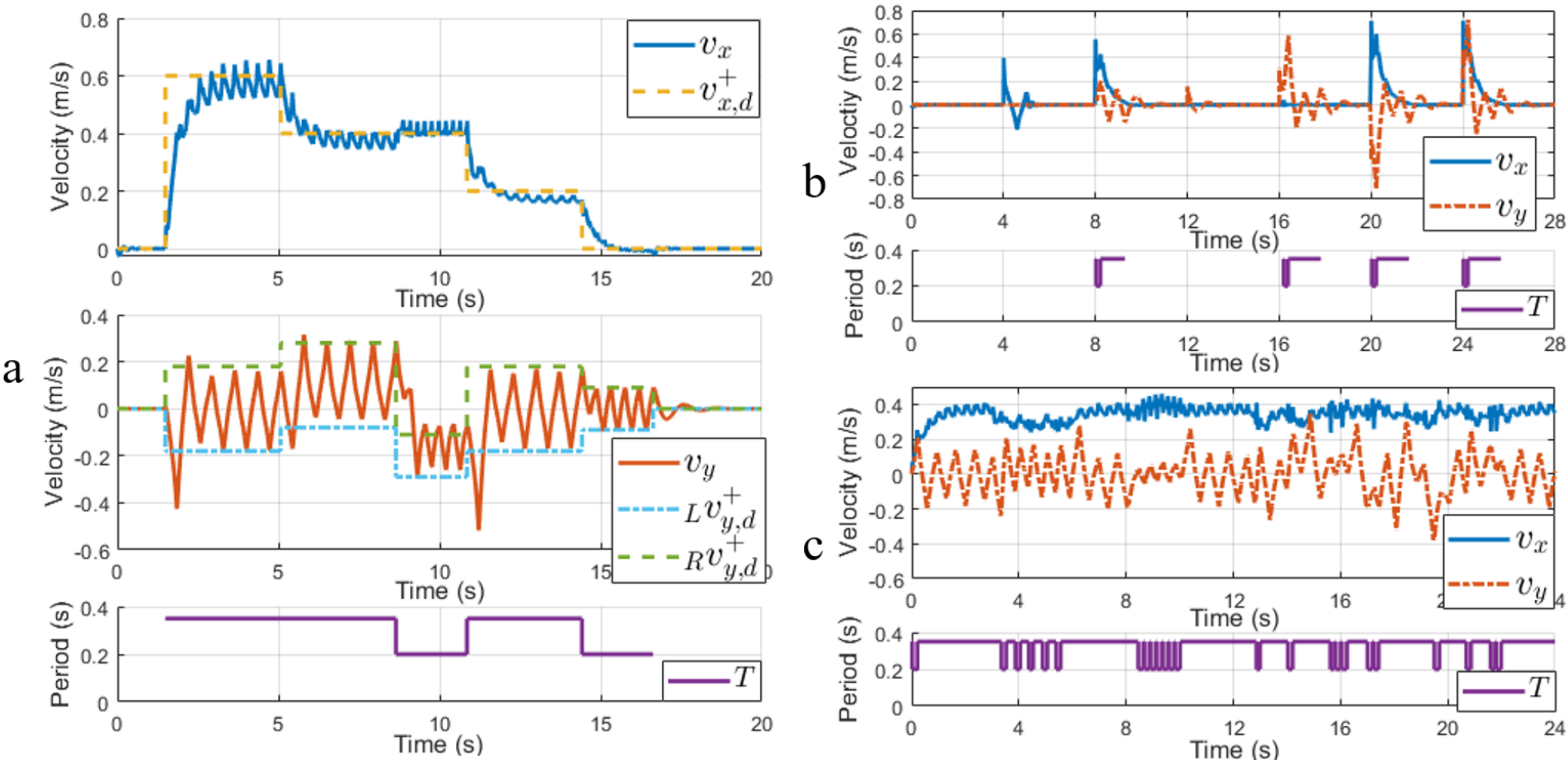}
   \caption{Plots of velocity and gait period of the robot for (a) versatile walking, (b) push recovery and (c) rough terrain walking in simulations. In the period plots, $T$ is $\infty$ in the blank segments, i.e., the robot is standing.}
      \label{pic:sim_cmd_walk}
   \end{center}
\end{figure}

The first simulation is walking and standing following the user command. As shown in Fig. \ref{pic:sim_cmd_walk}a, the robot started from standing, then transitioned to walking following the desired velocity and period commands, finally returned to standing. We can see that the robot followed the desired $v_{x,d}^+, {}_Rv_{y,d}^+,{}_Lv_{y,d}^+$ very responsively and accurately.
\begin{table}[htb]
    \caption{The time and magnitude of impulses.}\label{tab:impulses}
	\centering
	\resizebox{0.4\textwidth}{!}{$
	\begin{tabular}{c|c|c|c|c|c|c}
        \hline
		Time & 4s & 8s & 12s & 16s & 20s & 24s \\      
        \hline
        Sagittal & 5Ns & 7Ns & 0 & 0 & 9Ns & 9Ns \\
        \hline
        Lateral & 0 & 0 & 2Ns & 4Ns & -6Ns & 6Ns \\
        \hline
    \end{tabular}$}
\end{table}

The second simulation is the push-recovery test. The robot started from standing. Six impulses with different magnitudes were applied to the robot in the sagittal and lateral directions. The magnitude and applied time of these impulses are shown in Tab. \ref{tab:impulses}. As shown in Fig. \ref{pic:sim_cmd_walk}b, for the first and third impulses, the robot recovered to steady state purely by the standing controller. While for other cases, the robot detected that it can not keep standing and automatically took steps with appropriate periods and then returned to standing. The robot recovered from instant velocity change up to 0.7m/s in the sagittal direction and 0.5m/s in the lateral direction.

\begin{figure}[ht]
   \begin{center}
    \includegraphics[width=0.38\textwidth]{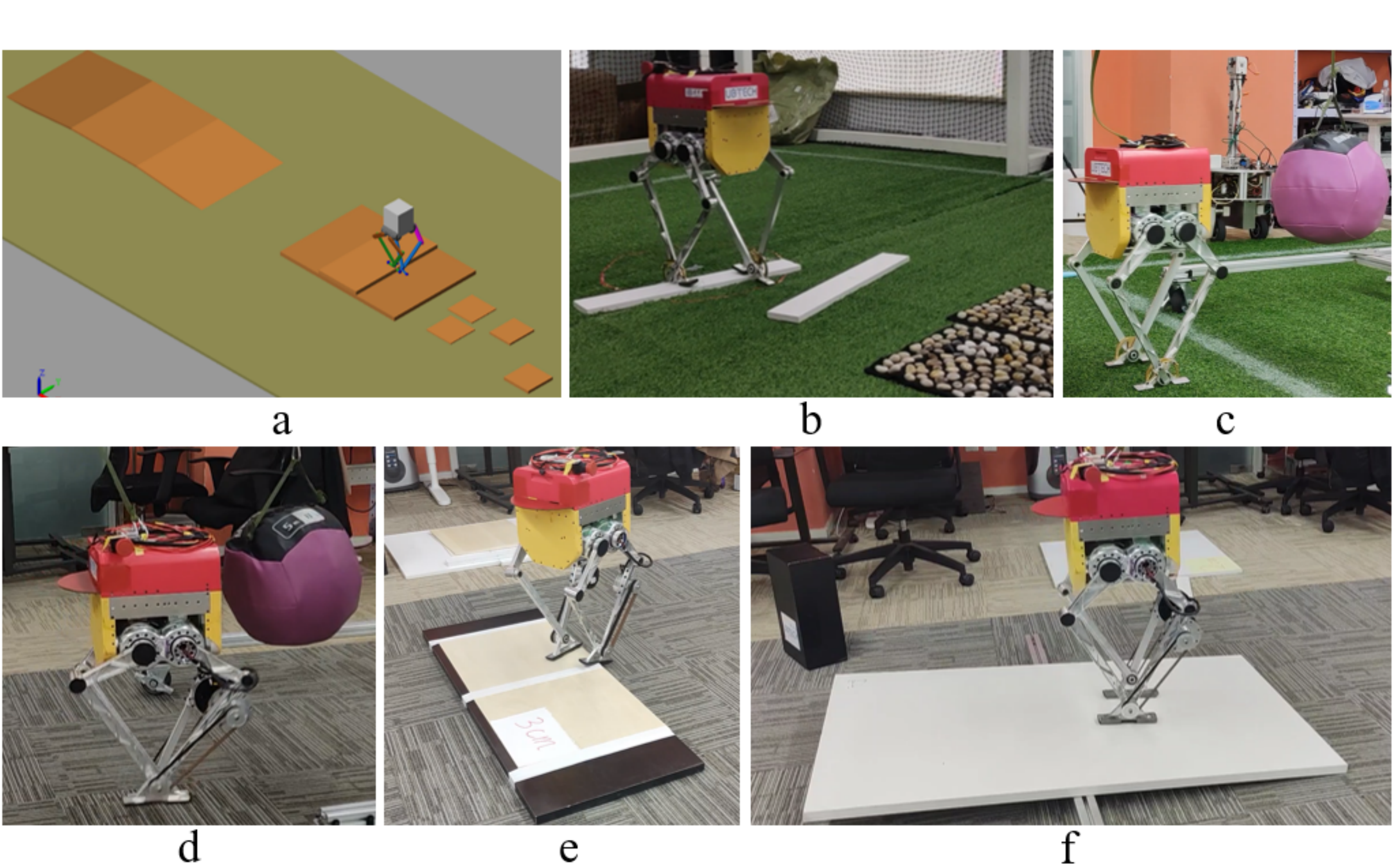}
   \caption{(a) shows the uneven terrain (15 degrees slopes, 5cm stairs and 2-5cm boards) in simulation. (b) shows the passive-foot robot walking over 2cm rough terrain. (c) and (d) show that the passive-foot and active-foot robot are hit by a 5kg wall ball during stepping in place. (e) shows the active-foot robot walking over a 3cm board. (f) shows the active-foot robot walking over a 5 degrees slope.}
      \label{pic:uneven_push}
   \end{center}
\end{figure}

The third simulation is the uneven terrain test. As shown in Fig. \ref{pic:uneven_push}a, the robot blindly walked over a terrain with 15 degree slopes, 5cm stairs (12\% of the hip height) and 2-5cm boards. The desired speed is 0.4m/s and the nominal period is 0.35s. The velocity and gait period are shown in Fig. \ref{pic:sim_cmd_walk}c, we can see that the robot successfully passed this terrain with small velocity variation.

\subsection{Experimental Results}
\begin{figure}[htb]
   \begin{center}
    \includegraphics[width=0.42\textwidth]{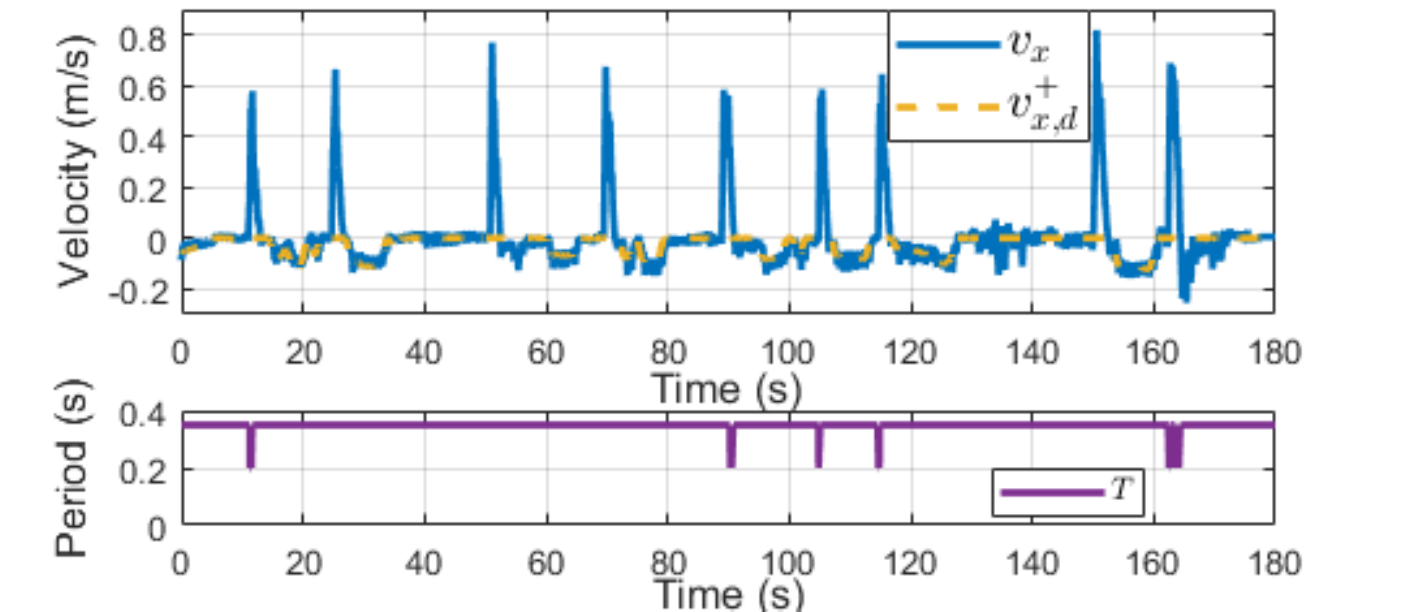}
   \caption{The first figure shows the sagittal velocity of the robot during the push recovery experiment. The second figure shows the gait period.}
      \label{pic:push_exp}
   \end{center}
\end{figure}

Finally we apply the proposed gait synthesizer to both the passive-foot and active-foot physical robots in Fig. \ref{pic:Mario}. The robot achieved stable walking with maximal speed 0.7m/s, passed the push-recovery tests and preliminary uneven-terrain tests as shown in Fig.\ref{pic:uneven_push}. The experimental data of the push-recovery test for the active-foot robot is presented here. As shown in Fig. \ref{pic:uneven_push}d, the robot was hit by a 5kg (29\% of its weight) wall ball in the sagittal direction for 9 times while stepping in place. The sagittal velocity and gait period are shown in Fig. \ref{pic:push_exp}, the robot recovered from instant velocity change up to 0.8m/s.

\begin{figure}[htb]
   \begin{center}
    \includegraphics[width=0.45\textwidth]{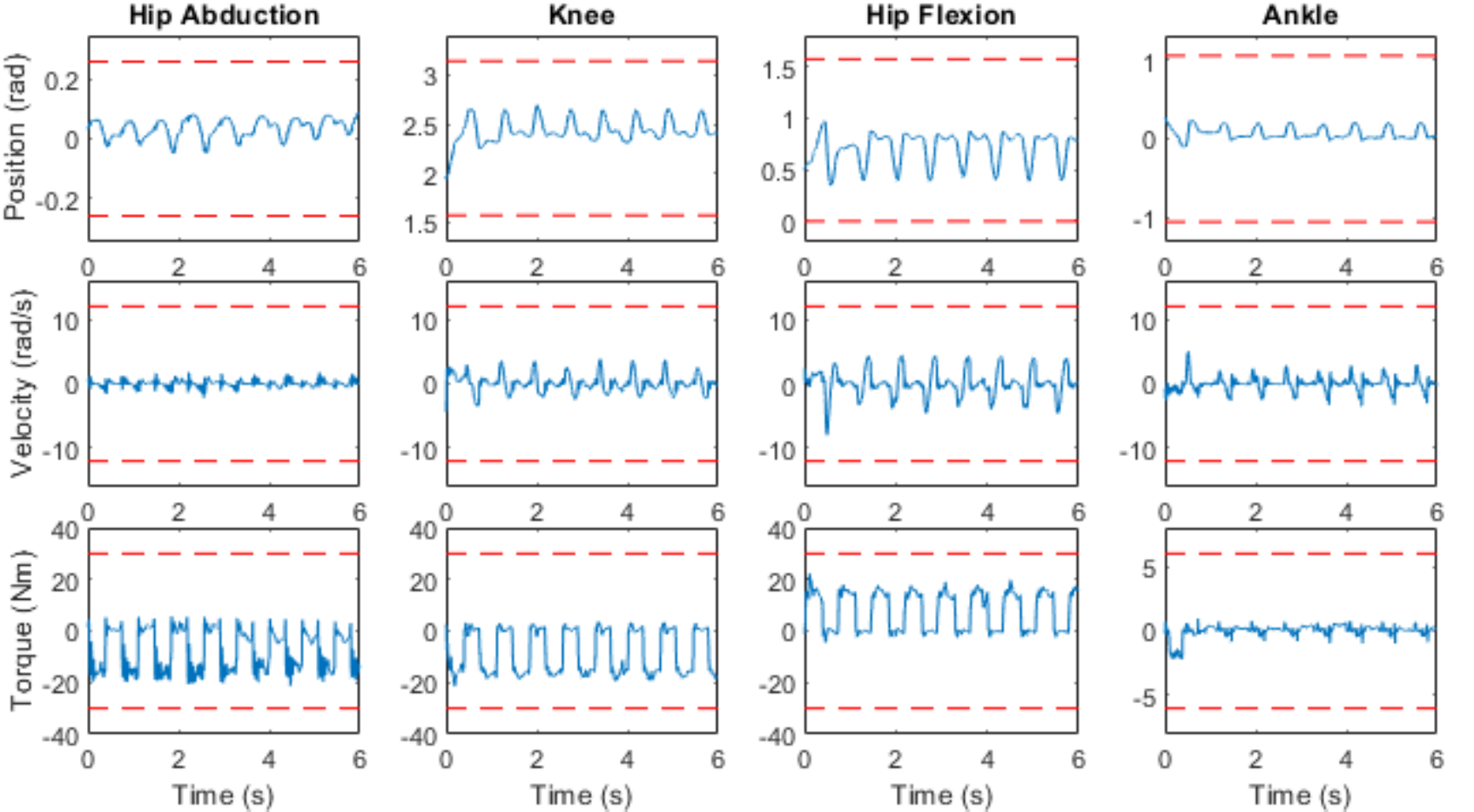}
   \caption{Experimental data plots of the joint positions, velocities and torques of the right leg after a push disturbance.}
      \label{pic:push_exp_joint}
   \end{center}
\end{figure}
\begin{figure}[htb]
   \begin{center}
    \includegraphics[width=0.5\textwidth]{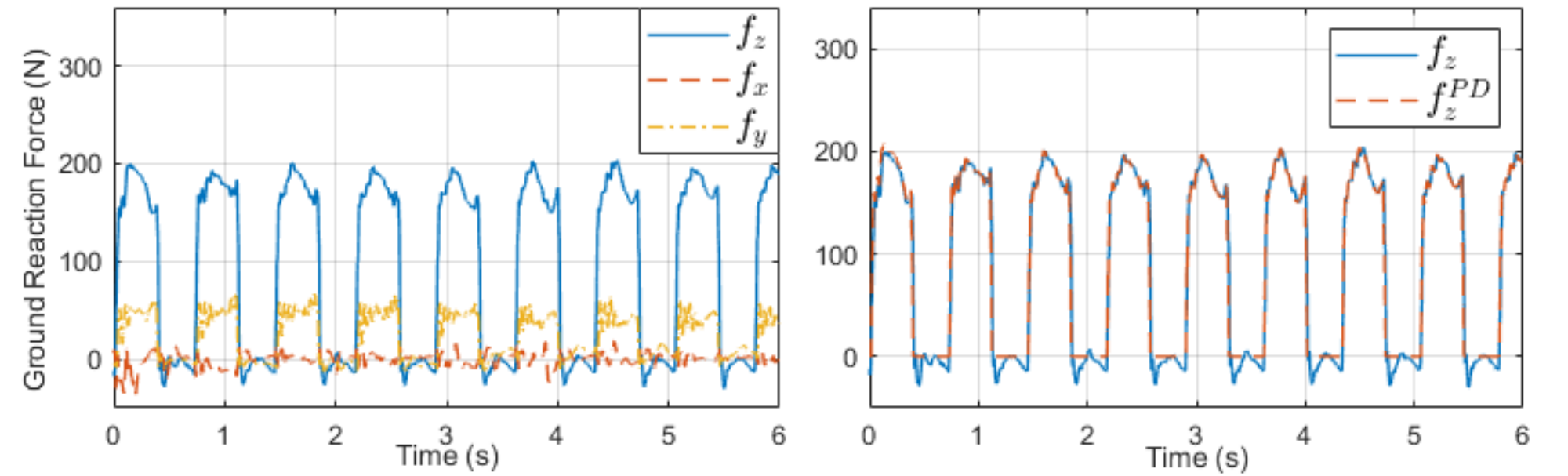}
   \caption{Experimental data plots of the right leg GRFs after a push disturbance, and the comparison of $f_z$ with the value of the PD law (\ref{eq:fz}).}
   \label{pic:push_exp_GRF}
   \end{center}
\end{figure}

The experimental data plots of the joints and ground reaction forces (estimated with joint torques) of the right leg just after a push disturbance are shown in Fig. \ref{pic:push_exp_joint} and Fig. \ref{pic:push_exp_GRF}. We can see that all joint positions, velocities and torques are well within the bounds represented by the dashed lines. The ground reaction forces also well satisfy the friction cone constraints ($\mu$=0.6). These results help to demonstrate the feasibility of the generated gaits. Furthermore, the estimated vertical ground reaction force $f_z$ is compared with the value from the PD law (\ref{eq:fz}), we can see they are almost identical during the stance phase, thus (\ref{eq:fz}) can used to predict $f_z$.

\section{CONCLUSION}\label{sec:conclusion}
The results in this paper provide a methodology towards full-dimensional model based real-time motion planning for bipedal locomotion with UUB stability. We showed that our proposed gait synthesizer can provide feasible solutions to the constructed MPC optimization problems, which leads to fast online planning at 1kHz. The proof of stability is provided by showing that the post-impact CoM states of the robot are UUB stable. Simulation and experimental results showed that, with this proposed approach, robots can achieve flexible transitions between standing and walking, accurate velocity tracking with different step periods, robust locomotion under disturbances, and passing uneven terrains. 

Future work will be extending current work to locomotion planning with terrain knowledge. Exploring more intelligent methods of combining whole-body and centroidal dynamics for motion planning and combining other methods of generating centroidal trajectories \cite{fernbach2020c} are also future directions.

\section*{ACKNOWLEDGMENT}
This work was supported by UBTECH Robotics and the Robotic Control Lab in Tsinghua University.

\bibliographystyle{IEEEtran}

\begin{thebibliography}{10}
\providecommand{\url}[1]{#1}
\csname url@rmstyle\endcsname
\providecommand{\newblock}{\relax}
\providecommand{\bibinfo}[2]{#2}
\providecommand\BIBentrySTDinterwordspacing{\spaceskip=0pt\relax}
\providecommand\BIBentryALTinterwordstretchfactor{4}
\providecommand\BIBentryALTinterwordspacing{\spaceskip=\fontdimen2\font plus
\BIBentryALTinterwordstretchfactor\fontdimen3\font minus
  \fontdimen4\font\relax}
\providecommand\BIBforeignlanguage[2]{{%
\expandafter\ifx\csname l@#1\endcsname\relax
\typeout{** WARNING: IEEEtran.bst: No hyphenation pattern has been}%
\typeout{** loaded for the language `#1'. Using the pattern for}%
\typeout{** the default language instead.}%
\else
\language=\csname l@#1\endcsname
\fi
#2}}

\bibitem{raibert1986legged}
M.~H. Raibert, \emph{Legged robots that balance}.\hskip 1em plus 0.5em minus
  0.4em\relax MIT press, 1986.

\bibitem{kajita2014introduction}
S.~Kajita, H.~Hirukawa, K.~Harada, and K.~Yokoi, \emph{Introduction to humanoid
  robotics}.\hskip 1em plus 0.5em minus 0.4em\relax Springer, 2014, vol. 101.

\bibitem{englsberger2015three}
J.~Englsberger, C.~Ott, and A.~Albu-Sch{\"a}ffer, ``Three-dimensional bipedal
  walking control based on divergent component of motion,'' \emph{IEEE
  Transactions on Robotics}, vol.~31, no.~2, pp. 355--368, 2015.

\bibitem{dai2014whole}
H.~Dai, A.~Valenzuela, and R.~Tedrake, ``Whole-body motion planning with
  centroidal dynamics and full kinematics,'' in \emph{2014 IEEE-RAS
  International Conference on Humanoid Robots (Humanoids)}.\hskip 1em plus
  0.5em minus 0.4em\relax IEEE, 2014, pp. 295--302.

\bibitem{posa2016optimization}
M.~Posa, S.~Kuindersma, and R.~Tedrake, ``Optimization and stabilization of
  trajectories for constrained dynamical systems,'' in \emph{2016 IEEE
  International Conference on Robotics and Automation (ICRA)}.\hskip 1em plus
  0.5em minus 0.4em\relax IEEE, 2016, pp. 1366--1373.

\bibitem{dafarra2020whole}
S.~Dafarra, G.~Romualdi, G.~Metta, and D.~Pucci, ``Whole-body walking
  generation using contact parametrization: A non-linear trajectory
  optimization approach,'' in \emph{2020 IEEE International Conference on
  Robotics and Automation (ICRA)}.\hskip 1em plus 0.5em minus 0.4em\relax IEEE,
  2020, pp. 1511--1517.

\bibitem{westervelt2003switching}
E.~R. Westervelt, J.~W. Grizzle, and C.~C. De~Wit, ``Switching and pi control
  of walking motions of planar biped walkers,'' \emph{IEEE Transactions on
  Automatic Control}, vol.~48, no.~2, pp. 308--312, 2003.

\bibitem{powell2013speed}
M.~J. Powell, A.~Hereid, and A.~D. Ames, ``Speed regulation in 3d robotic
  walking through motion transitions between human-inspired partial hybrid zero
  dynamics,'' in \emph{2013 IEEE International Conference on Robotics and
  Automation (ICRA)}.\hskip 1em plus 0.5em minus 0.4em\relax IEEE, 2013, pp.
  4803--4810.

\bibitem{murali2019optimal}
V.~Murali, A.~D. Ames, and E.~I. Verriest, ``Optimal walking speed transitions
  for fully actuated bipedal robots,'' in \emph{2019 IEEE 58th Conference on
  Decision and Control (CDC)}.\hskip 1em plus 0.5em minus 0.4em\relax IEEE,
  2019, pp. 6295--6300.

\bibitem{da20162d}
X.~Da, O.~Harib, R.~Hartley, B.~Griffin, and J.~W. Grizzle, ``From 2d design of
  underactuated bipedal gaits to 3d implementation: Walking with speed
  tracking,'' \emph{IEEE Access}, vol.~4, pp. 3469--3478, 2016.

\bibitem{feng2016online}
S.~Feng, ``Online hierarchical optimization for humanoid control,'' Ph.D.
  dissertation, Carnegie Mellon University, 2016.

\bibitem{koolen2012capturability}
T.~Koolen, T.~De~Boer, J.~Rebula, A.~Goswami, and J.~Pratt,
  ``Capturability-based analysis and control of legged locomotion, part 1:
  Theory and application to three simple gait models,'' \emph{The International
  Journal of Robotics Research}, vol.~31, no.~9, pp. 1094--1113, 2012.

\bibitem{wang2017robust}
H.~Wang and M.~Zhao, ``A robust biped gait controller using step timing
  optimization with fixed footprint constraints,'' in \emph{2017 IEEE
  International Conference on Robotics and Biomimetics (ROBIO)}.\hskip 1em plus
  0.5em minus 0.4em\relax IEEE, 2017, pp. 1787--1792.

\bibitem{khadiv2020walking}
M.~Khadiv, A.~Herzog, S.~A.~A. Moosavian, and L.~Righetti, ``Walking control
  based on step timing adaptation,'' \emph{IEEE Transactions on Robotics},
  2020.

\bibitem{scianca2020mpc}
N.~Scianca, D.~De~Simone, L.~Lanari, and G.~Oriolo, ``Mpc for humanoid gait
  generation: Stability and feasibility,'' \emph{IEEE Transactions on
  Robotics}, vol.~36, no.~4, pp. 1171--1188, 2020.

\bibitem{smaldone2021feasibility}
F.~M. Smaldone, N.~Scianca, L.~Lanari, and G.~Oriolo, ``Feasibility-driven step
  timing adaptation for robust mpc-based gait generation in humanoids,''
  \emph{IEEE Robotics and Automation Letters}, vol.~6, no.~2, pp. 1582--1589,
  2021.

\bibitem{daneshmand2021variable}
E.~Daneshmand, M.~Khadiv, F.~Grimminger, and L.~Righetti, ``Variable horizon
  mpc with swing foot dynamics for bipedal walking control,'' \emph{IEEE
  Robotics and Automation Letters}, vol.~6, no.~2, pp. 2349--2356, 2021.

\bibitem{westervelt2018feedback}
E.~R. Westervelt, J.~W. Grizzle, C.~Chevallereau, J.~H. Choi, and B.~Morris,
  \emph{Feedback control of dynamic bipedal robot locomotion}.\hskip 1em plus
  0.5em minus 0.4em\relax CRC press, 2018.

\bibitem{gong2019feedback}
Y.~Gong, R.~Hartley, X.~Da, A.~Hereid, O.~Harib, J.-K. Huang, and J.~Grizzle,
  ``Feedback control of a cassie bipedal robot: Walking, standing, and riding a
  segway,'' in \emph{2019 American Control Conference (ACC)}.\hskip 1em plus
  0.5em minus 0.4em\relax IEEE, 2019, pp. 4559--4566.

\bibitem{hereid2017frost}
A.~Hereid and A.~D. Ames, ``Frost: Fast robot optimization and simulation
  toolkit,'' in \emph{2017 IEEE/RSJ International Conference on Intelligent
  Robots and Systems (IROS)}.\hskip 1em plus 0.5em minus 0.4em\relax IEEE,
  2017, pp. 719--726.

\bibitem{hereid2019rapid}
A.~Hereid, O.~Harib, R.~Hartley, Y.~Gong, and J.~W. Grizzle, ``Rapid trajectory
  optimization using c-frost with illustration on a cassie-series dynamic
  walking biped,'' in \emph{2019 IEEE/RSJ International Conference on
  Intelligent Robots and Systems (IROS)}.\hskip 1em plus 0.5em minus
  0.4em\relax IEEE, 2019, pp. 4722--4729.

\bibitem{xiong2019orbit}
X.~Xiong and A.~D. Ames, ``Orbit characterization, stabilization and
  composition on 3d underactuated bipedal walking via hybrid passive linear
  inverted pendulum model,'' in \emph{2019 IEEE/RSJ International Conference on
  Intelligent Robots and Systems (IROS)}.\hskip 1em plus 0.5em minus
  0.4em\relax IEEE, 2019, pp. 4644--4651.

\bibitem{koolen2016balance}
T.~Koolen, M.~Posa, and R.~Tedrake, ``Balance control using center of mass
  height variation: limitations imposed by unilateral contact,'' in \emph{2016
  IEEE-RAS 16th International Conference on Humanoid Robots (Humanoids)}.\hskip
  1em plus 0.5em minus 0.4em\relax IEEE, 2016, pp. 8--15.

\bibitem{rezazadeh2015spring}
S.~Rezazadeh, C.~Hubicki, M.~Jones, A.~Peekema, J.~Van~Why, A.~Abate, and
  J.~Hurst, ``Spring-mass walking with atrias in 3d: Robust gait control
  spanning zero to 4.3 kph on a heavily underactuated bipedal robot,'' in
  \emph{2015 ASME Dynamic Systems and Control Conference}.\hskip 1em plus 0.5em
  minus 0.4em\relax ASME, 2015.

\bibitem{apgar2018fast}
T.~Apgar, P.~Clary, K.~Green, A.~Fern, and J.~W. Hurst, ``Fast online
  trajectory optimization for the bipedal robot cassie.'' in \emph{Robotics:
  Science and Systems}, vol. 101, 2018, p.~14.

\bibitem{fernbach2020c}
P.~Fernbach, S.~Tonneau, O.~Stasse, J.~Carpentier, and M.~Ta{\"\i}x, ``C-croc:
  Continuous and convex resolution of centroidal dynamic trajectories for
  legged robots in multicontact scenarios,'' \emph{IEEE Transactions on
  Robotics}, vol.~36, no.~3, pp. 676--691, 2020.

\end{thebibliography}

%%%%%%%%%%%%%%%%%%%%%%%%%%%%%%%%%%%%%%%%%%%%%%%%%%%%%%%%%%%%%%%%%%%%%%%%%%%%%%%%

\end{document}